\documentclass[pmlr,twocolumn,10pt]{jmlr}

\mlhtrack{proceedings}

\newif\iffinal
\finaltrue
\newcommand{\CameraReadyPMLRVolume}{346}

\iffinal
  \ifx\CameraReadyPMLRVolume\empty
    \jmlrproceedings{}{ML4H 2026: Proceedings Track}
  \else
    \jmlrvolume{\CameraReadyPMLRVolume}
  \fi
  \jmlryear{2026}
  \jmlrworkshop{Machine Learning for Health (ML4H) 2026}
\else
  \jmlrproceedings{}{Submitted to ML4H 2026: \mlhtrackname}
  \jmlrworkshop{Machine Learning for Health (ML4H) 2026}
  \makeatletter
  \renewcommand*{\@titlefoot}{}
  \makeatother
\fi

\usepackage{array}
\usepackage{booktabs}
\usepackage{float}
\usepackage[nameinlink,noabbrev]{cleveref}
\crefname{figure}{Figure}{Figures}
\crefname{table}{Table}{Tables}
\usepackage{microtype}
\usepackage{placeins}
\usepackage{pgfplots}
\usepackage{tikz}
\usepackage{xspace}
\usepackage[switch]{lineno}
\usetikzlibrary{arrows.meta,positioning}
\pgfplotsset{compat=1.18}
\hypersetup{
  hypertexnames=false,
  pdftitle={PatientFlowSynth: Evaluating FHIR Agents with Matched Patient Charts},
  pdfauthor={Anthony Cuturrufo, Ratul Ghosh, Devashish Khatwani, Krishnaram Kenthapadi},
  pdfsubject={ML4H 2026 Proceedings Track},
  pdfkeywords={FHIR, medical AI agents, matched chart variants, evidence grounding}
}


\definecolor{pfBlue}{HTML}{0072B2}
\definecolor{pfTeal}{HTML}{009E73}
\definecolor{pfVermillion}{HTML}{D55E00}
\definecolor{pfOrange}{HTML}{E69F00}
\definecolor{pfPurple}{HTML}{8C5AA6}
\definecolor{pfGray}{HTML}{6B7280}
\definecolor{pfInk}{HTML}{24303F}
\definecolor{pfLight}{HTML}{F5F7FA}
\newcommand{\system}{\textsc{PatientFlowSynth}\xspace}

\newcommand{\modelLuna}{GPT-5.6 Luna\xspace}
\newcommand{\modelTerra}{GPT-5.6 Terra\xspace}
\newcommand{\supported}{\texttt{supported}\xspace}
\newcommand{\refuted}{\texttt{refuted}\xspace}
\newcommand{\insufficient}{\texttt{insufficient\_evidence}\xspace}
\newcommand{\conflicting}{\texttt{conflicting\_evidence}\xspace}
\newcolumntype{L}[1]{>{\raggedright\arraybackslash}p{#1}}
\newcommand{\PFSReleasePatients}{225}

\newcommand{\PFSReleaseCells}{7,200}
\newcommand{\PFSReleaseFamilies}{900}
\newcommand{\PFSReleaseTasks}{18,000}

\newcommand{\PFSSplitTrainTasks}{12,560}
\newcommand{\PFSSplitTrainPatients}{157}
\newcommand{\PFSSplitDevTasks}{2,720}
\newcommand{\PFSSplitDevPatients}{34}
\newcommand{\PFSSplitTestTasks}{2,720}
\newcommand{\PFSSplitTestPatients}{34}
\newcommand{\PFSConfirmPatients}{225}
\newcommand{\PFSConfirmPairs}{3,600}
\newcommand{\PFSConfirmEpisodes}{7,200}
\newcommand{\PFSConfirmPairsPerPatient}{16}
\newcommand{\PFSConfirmBootstrapDraws}{50,000}
\newcommand{\PFSConfirmSignFlipDraws}{100,000}
\newcommand{\PFSConfirmSignFlipDenominator}{100{,}001}
\newcommand{\PFSConfirmStrictNaturalOne}{40.7}
\newcommand{\PFSConfirmStrictNaturalTwo}{40.72}
\newcommand{\PFSConfirmStrictExplicitOne}{51.3}
\newcommand{\PFSConfirmStrictExplicitTwo}{51.28}

\newcommand{\PFSConfirmDTwoWOneNatural}{48.89}
\newcommand{\PFSConfirmDTwoWOneExplicit}{53.33}
\newcommand{\PFSConfirmDTwoWOneDiff}{4.44}

\newcommand{\PFSConfirmDTwoWTwoNatural}{32.56}
\newcommand{\PFSConfirmDTwoWTwoExplicit}{49.22}
\newcommand{\PFSConfirmDTwoWTwoDiff}{16.67}
\newcommand{\PFSDepthPatients}{225}
\newcommand{\PFSDepthPairsPerTopology}{1,800}
\newcommand{\PFSDepthDThreeWOneNatural}{8.94}
\newcommand{\PFSDepthDThreeWOneExplicit}{13.06}
\newcommand{\PFSDepthDThreeWOneDiff}{4.11}

\newcommand{\PFSDepthDThreeWTwoNatural}{4.11}
\newcommand{\PFSDepthDThreeWTwoExplicit}{7.28}
\newcommand{\PFSDepthDThreeWTwoDiff}{3.17}

\newcommand{\PFSConfirmStrictDiffOne}{10.6}
\newcommand{\PFSConfirmStrictDiffTwo}{10.56}
\newcommand{\PFSConfirmStrictCILowOne}{8.4}
\newcommand{\PFSConfirmStrictCILowTwo}{8.36}
\newcommand{\PFSConfirmStrictCIHighOne}{12.8}
\newcommand{\PFSConfirmStrictCIHighTwo}{12.75}
\newcommand{\PFSConfirmPositivePatients}{155}
\newcommand{\PFSConfirmNegativePatients}{54}
\newcommand{\PFSConfirmZeroPatients}{16}
\newcommand{\PFSConfirmSensitivityMin}{10.49}
\newcommand{\PFSConfirmSensitivityMax}{10.87}
\newcommand{\PFSConfirmSensitivityCILowMin}{8.28}

\newcommand{\PFSMetricExactNatural}{61.50}
\newcommand{\PFSMetricExactExplicit}{73.00}
\newcommand{\PFSMetricExactDiff}{11.50}
\newcommand{\PFSMetricExactCILow}{9.69}
\newcommand{\PFSMetricExactCIHigh}{13.28}
\newcommand{\PFSMetricEvidenceNatural}{68.17}
\newcommand{\PFSMetricEvidenceExplicit}{93.78}
\newcommand{\PFSMetricEvidenceDiff}{25.61}
\newcommand{\PFSMetricEvidenceCILow}{23.64}
\newcommand{\PFSMetricEvidenceCIHigh}{27.56}
\newcommand{\PFSMetricPermissiveNatural}{52.64}
\newcommand{\PFSMetricPermissiveExplicit}{69.89}
\newcommand{\PFSMetricPermissiveDiff}{17.25}
\newcommand{\PFSMetricPermissiveCILow}{15.33}
\newcommand{\PFSMetricPermissiveCIHigh}{19.19}
\newcommand{\PFSMetricSearchNatural}{40.75}
\newcommand{\PFSMetricSearchExplicit}{51.31}
\newcommand{\PFSMetricSearchDiff}{10.56}
\newcommand{\PFSMetricSearchCILow}{8.36}
\newcommand{\PFSMetricSearchCIHigh}{12.78}
\newcommand{\PFSMetricStrictNatural}{40.72}
\newcommand{\PFSMetricStrictExplicit}{51.28}
\newcommand{\PFSMetricStrictDiff}{10.56}
\newcommand{\PFSMetricStrictCILow}{8.36}
\newcommand{\PFSMetricStrictCIHigh}{12.75}

\newcommand{\PFSMissingEvidenceNatural}{71.44}
\newcommand{\PFSMissingStrictNatural}{18.56}
\newcommand{\PFSMissingPrimaryStateNatural}{49.22}

\newcommand{\PFSMissingEvidenceExplicit}{97.00}
\newcommand{\PFSMissingStrictExplicit}{24.56}
\newcommand{\PFSMissingPrimaryStateExplicit}{29.33}
\newcommand{\PFSMissingEpisodesPerCell}{900}

\newcommand{\PFSMultiDefectEpisodes}{2,048}

\newcommand{\PFSMultiDefectStrictDiff}{6.84}
\newcommand{\PFSMultiDefectStrictCILow}{2.73}
\newcommand{\PFSMultiDefectStrictCIHigh}{10.94}
\newcommand{\PFSMultiDefectStrictP}{0.00148}

\newcommand{\PFSMultiDefectEncounterLunaPlacebo}{38.28}
\newcommand{\PFSMultiDefectEncounterLunaRelevant}{50.78}
\newcommand{\PFSMultiDefectEncounterLunaDiff}{12.50}
\newcommand{\PFSMultiDefectEncounterLunaCILow}{1.56}
\newcommand{\PFSMultiDefectEncounterLunaCIHigh}{23.44}
\newcommand{\PFSMultiDefectEncounterTerraPlacebo}{43.75}
\newcommand{\PFSMultiDefectEncounterTerraRelevant}{77.34}
\newcommand{\PFSMultiDefectEncounterTerraDiff}{33.59}
\newcommand{\PFSMultiDefectEncounterTerraCILow}{25.78}
\newcommand{\PFSMultiDefectEncounterTerraCIHigh}{41.41}
\newcommand{\PFSMultiDefectConditionLunaPlacebo}{42.97}
\newcommand{\PFSMultiDefectConditionLunaRelevant}{53.91}
\newcommand{\PFSMultiDefectConditionLunaDiff}{10.94}
\newcommand{\PFSMultiDefectConditionLunaCILow}{1.56}
\newcommand{\PFSMultiDefectConditionLunaCIHigh}{20.31}
\newcommand{\PFSMultiDefectConditionTerraPlacebo}{30.47}
\newcommand{\PFSMultiDefectConditionTerraRelevant}{24.22}
\newcommand{\PFSMultiDefectConditionTerraDiff}{-6.25}
\newcommand{\PFSMultiDefectConditionTerraCILow}{-16.41}
\newcommand{\PFSMultiDefectConditionTerraCIHigh}{3.13}
\newcommand{\PFSMultiDefectImmunizationLunaPlacebo}{33.59}
\newcommand{\PFSMultiDefectImmunizationLunaRelevant}{38.28}
\newcommand{\PFSMultiDefectImmunizationLunaDiff}{4.69}
\newcommand{\PFSMultiDefectImmunizationLunaCILow}{-5.47}
\newcommand{\PFSMultiDefectImmunizationLunaCIHigh}{14.84}
\newcommand{\PFSMultiDefectImmunizationTerraPlacebo}{14.84}
\newcommand{\PFSMultiDefectImmunizationTerraRelevant}{25.00}
\newcommand{\PFSMultiDefectImmunizationTerraDiff}{10.16}
\newcommand{\PFSMultiDefectImmunizationTerraCILow}{1.56}
\newcommand{\PFSMultiDefectImmunizationTerraCIHigh}{18.75}
\newcommand{\PFSMultiDefectObservationLunaPlacebo}{44.53}
\newcommand{\PFSMultiDefectObservationLunaRelevant}{42.19}
\newcommand{\PFSMultiDefectObservationLunaDiff}{-2.34}
\newcommand{\PFSMultiDefectObservationLunaCILow}{-17.19}
\newcommand{\PFSMultiDefectObservationLunaCIHigh}{11.72}
\newcommand{\PFSMultiDefectObservationTerraPlacebo}{52.34}
\newcommand{\PFSMultiDefectObservationTerraRelevant}{43.75}
\newcommand{\PFSMultiDefectObservationTerraDiff}{-8.59}
\newcommand{\PFSMultiDefectObservationTerraCILow}{-19.53}
\newcommand{\PFSMultiDefectObservationTerraCIHigh}{2.34}

\newcommand{\PFSTraceObservationLunaMissingSupportedPlacebo}{3}
\newcommand{\PFSTraceObservationLunaMissingSupportedRelevant}{15}
\newcommand{\PFSTraceObservationTerraConflictSupportedPlacebo}{7}
\newcommand{\PFSTraceObservationTerraConflictSupportedRelevant}{18}
\newcommand{\PFSTraceEncounterTerraMissingRetrievedPlacebo}{16}
\newcommand{\PFSTraceEncounterTerraMissingRetrievedRelevant}{32}
\newcommand{\PFSTraceEncounterTerraMissingExactPlacebo}{25}
\newcommand{\PFSTraceEncounterTerraMissingExactRelevant}{10}
\newcommand{\PFSTraceEncounterTerraMissingStrictPlacebo}{9}

\newcommand{\PFSReplicationPatients}{48}
\newcommand{\PFSReplicationEpisodes}{6,144}
\newcommand{\PFSReplicationLunaNatural}{37.76}
\newcommand{\PFSReplicationLunaExplicit}{48.89}
\newcommand{\PFSReplicationLunaDiff}{11.13}
\newcommand{\PFSReplicationLunaCILow}{6.84}
\newcommand{\PFSReplicationLunaCIHigh}{15.43}
\newcommand{\PFSReplicationTerraNatural}{35.87}
\newcommand{\PFSReplicationTerraExplicit}{80.40}
\newcommand{\PFSReplicationTerraDiff}{44.53}
\newcommand{\PFSReplicationTerraCILow}{41.21}
\newcommand{\PFSReplicationTerraCIHigh}{47.79}

\newcommand{\PFSTemplateSignatureCount}{9}
\newcommand{\PFSTemplatePositiveSignatureCount}{8}

\newcommand{\PFSFamilyAuditItems}{1,536}
\providecommand{\Description}[1]{}

\title[PatientFlowSynth: Evaluating FHIR Agents]
{PatientFlowSynth: Evaluating FHIR Agents\\with Matched Patient Charts}

\author[Cuturrufo et al.]{%
\Name{Anthony Cuturrufo} \Email{anthony.cuturrufo@oracle.com}\\
\Name{Ratul Ghosh} \Email{ratul.ghosh@oracle.com}\\
\Name{Devashish Khatwani} \Email{devashish.khatwani@oracle.com}\\
\Name{Krishnaram Kenthapadi} \Email{krishnaram.kenthapadi@oracle.com}\\
\addr Oracle Health AI
}

\iffinal\else\linenumbers\fi
\begin{document}

\maketitle

\begin{abstract}
Agents that search patient charts must classify claims from retrieved records, including cases
with missing or conflicting evidence. In Fast Healthcare Interoperability Resources (FHIR),
status fields can change how those records should be interpreted. \system provides
\PFSReleaseTasks{} tasks from 225 synthetic patients and 18 templates that hold the patient, question, and retrieval program fixed across
chart variants representing four evidence states. Re-executing the program verifies each variant's
reference answer, evidence sets, and search requirements. Strict success requires a completed
response with correct labels and citations exactly matching an accepted evidence set retrieved
during the episode. Where required, a tool response must also show a complete search at the
specified scope. In a study using
\modelLuna{} and \PFSConfirmPatients{} synthetic patients (\PFSConfirmPairs{} matched pairs),
explicit Encounter-status guidance increased strict success from \PFSConfirmStrictNaturalOne\% to
\PFSConfirmStrictExplicitOne\% (+\PFSConfirmStrictDiffOne{} percentage points; 95\% patient-bootstrap
confidence interval, \PFSConfirmStrictCILowOne{} to \PFSConfirmStrictCIHighOne).
Among cases whose primary reference label was insufficient evidence, label agreement fell from
\pgfmathprintnumber[fixed,precision=1]{\PFSMissingPrimaryStateNatural}\% to
\pgfmathprintnumber[fixed,precision=1]{\PFSMissingPrimaryStateExplicit}\%.
A prespecified two-model replication reproduced both changes. Failed retrieval can mask
interpretation errors by producing a correct insufficient-evidence label without examining the
relevant records. Additional model comparisons showed that the benefit of guidance was
model-dependent. A visible-requirement external evaluation used 74 patients, and two clinical
reviewers independently agreed on 22 of 24 selected cases. Retrieval, citation, and classification
should be evaluated separately.
\end{abstract}


\ifmlhneedsstatements
\paragraph*{Data and Code Availability}
The public repository provides code for benchmark construction, verification, and evaluation,
plus Synthea generation instructions: \url{https://github.com/anthony-c-cuturrufo/PatientFlowSynth}.
The \href{https://github.com/anthony-c-cuturrufo/PatientFlowSynth/releases/tag/data-v1.0.0}{versioned training/development dataset}
contains 191 synthetic patients and 15,280 tasks from the primary depth-one/two benchmark. The 34-patient test set
remains private. The external evaluation uses the openly available, deidentified
MIMIC-IV-on-FHIR demo v2.0 from PhysioNet
\citep{bennett2022mimicfhirdemo,bennett2023mimicfhir,pollard2026physionet}; these source records are not redistributed.

\paragraph*{Institutional Review Board (IRB)}
The study used synthetic records and secondary analyses of openly available deidentified records,
with no patient interaction or intervention. Clinical reviewers assessed synthetic cases.
The authors assessed these activities as expert review and secondary data analysis that did not
require IRB review. No institutional or IRB determination was obtained.

\paragraph*{AI Assistance Disclosure}
Generative AI tools assisted with code development and language editing. The authors reviewed the
code, analyses, citations, and manuscript and take responsibility for the work.

\paragraph*{Funding and Competing Interests}
This work was conducted during an internship at Oracle Health AI, where all authors were
employed.
\fi

\section{Introduction}

A chart-search agent may be asked whether a patient received a vaccine during the most recent
completed visit. Here, the agent reads one patient's chart through read-only tools and returns
JSON labels for each claim and, when requested, their conjunction: \supported, \refuted,
\insufficient, or \conflicting. It also cites the record IDs that justify the labels.
The task evaluates retrieval and evidence classification, not treatment recommendations.

The Fast Healthcare Interoperability Resources (FHIR) standard represents clinical information as
typed, linked records. Correct retrieval does not guarantee correct interpretation: a
MedicationRequest describes a request, such as an order, and does not establish administration;
an Observation marked \texttt{entered-in-error} is not a negative result; and FHIR R4 encodes a
completed Encounter as \texttt{status=finished}; \texttt{completed} is not a valid Encounter status
\citep{hl7medicationrequestr4,hl7medicationadministrationr4,hl7observationr4,hl7encounterr4}.
An agent can retrieve the intended resources yet misread their lifecycle states, combine evidence
incorrectly, cite too few records, or claim absence after an incomplete search.
Answer accuracy alone does not distinguish these failures.

Clinical question-answering resources and medical-agent benchmarks evaluate query prediction,
FHIR retrieval, interactive workflows, or grounding in textual evidence
\citep{soni2020using,ryu2024ehrseqsql,frew2026fhirpathqa,
lee2026fhiragentbench,jiang2025medagentbench,soni2025archehrqa}.
\system compares agents across four chart-evidence states while holding the question and
executable decision rule fixed.

\paragraph{Contributions.}
First, we introduce \PFSReleaseTasks{} replay-validated FHIR tasks from 225 patients and 18 templates in matched families and use
fault injection to show why per-task replay alone does not validate a matched comparison.
Second, separate scores identify incorrect labels, insufficient or unobserved citations, and
invalid required searches; strict success also requires citations to match a declared evidence set
exactly. Third, Encounter-status guidance improved strict success but reduced agreement with the
program-derived \insufficient{} label in the primary study and a two-model replication.
This Encounter experiment is a focused
demonstration, not the full scope of the benchmark. Additional rule-specific, external, clinical,
and cross-model evaluations test its scoring and reveal limits to generalization.

\section{Related Work}

\paragraph{Executable Agent Evaluation.}
APIGen constructs function-calling data with format, execution, and semantic checks
\citep{liu2024apigen}. ToolSandbox evaluates stateful tool use through intermediate and
final milestones rather than requiring one fixed trajectory \citep{lu2025toolsandbox}.
MedAgentGym provides executable biomedical coding tasks and an environment for agent
training \citep{xu2025medagentgym}. These precedents establish execution-grounded
construction and evaluation. MedicalHarness further shows that medical-agent outcomes
depend on the model and its surrounding execution loop \citep{wang2026medicalharness}.
Our interface comparisons therefore hold the chart, tools, and task program fixed and
report retrieval, classification, citation, and strict-success outcomes separately.

\paragraph{EHR Question Answering and FHIR Workflows.}
EHRSQL includes hospital-derived questions and unanswerable requests, while EHR-SeqSQL
studies sequential, contextual questions and compositional generalization
\citep{lee2022ehrsql,ryu2024ehrseqsql}. FHIR-AgentBench evaluates clinical questions grounded
in interoperable records and compares retrieval and interaction strategies
\citep{lee2026fhiragentbench}. FHIR-AgentEval instead includes appointment-management and
genetic-testing workflows, with deterministic checks of agent responses and resulting
FHIR-server state \citep{mokssit2026fhiragenteval}. SCARE jointly evaluates question
answerability and verification or correction of candidate SQL \citep{lee2026scare}.
\system complements these settings with read-only chart searches and explicit
lifecycle-aware rules for supported, refuted, insufficient, and conflicting evidence.

\paragraph{Controlled Contrasts and Evidence-Grounded Verification.}
FEVER couples claim labels with supporting textual evidence, and contrast sets probe
models with small, meaning-changing edits \citep{thorne2018fever,gardner2020contrastsets}.
EHR-RobustGym extends paired evaluation to clinical agents with clean/noise questions and
SQL/Python interaction \citep{qiao2026ehrrobustgym}. Its construction changes a necessary
question condition while leaving the database unchanged. In contrast, \system fixes both
the question and its executable program, then changes permitted chart records to construct
four matched evidence states. Replay checks labels, sufficient evidence, and required
searches, while family-level checks verify the intended relation among chart variants.
The contribution is this construction and validation process, not verification labels or
controlled contrasts alone.

\paragraph{Scope of Executable Evidence Checks.}
Typed evidence certificates can gate agent termination \citep{liu2026evidencecarrying} or
validate EHR-study workflows \citep{fan2026procare}. Here, executable rules define a controlled
benchmark, not a learned stopping policy or a demonstration of training benefits.

\section{Task Definition and Evaluation Setting}

\label{sec:applied-setting}

% BEGIN inlined from figures/task_anatomy.tex
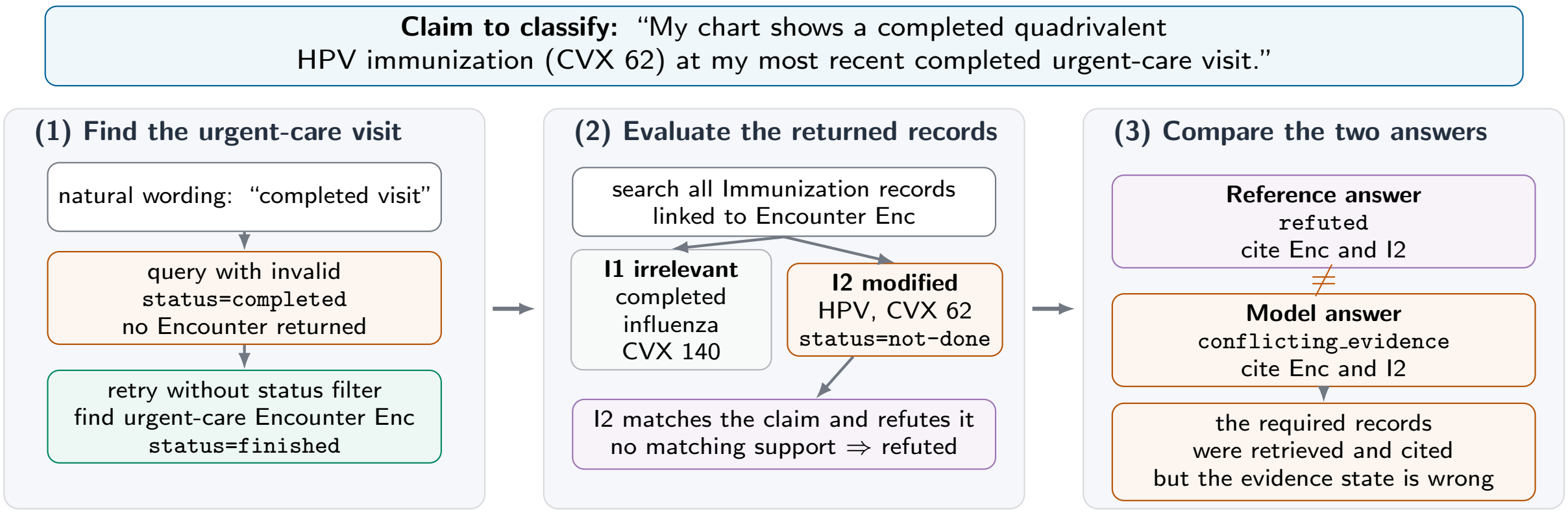
\begin{figure*}[t]
\centering
\begin{tikzpicture}[
  x=1cm,
  y=1cm,
  font=\sffamily\scriptsize,
  panel/.style={draw=pfInk!18,fill=pfLight,rounded corners=1.5mm},
  box/.style={draw=pfInk!65,fill=white,rounded corners=1.2mm,align=center,
    inner sep=3pt,minimum height=7mm},
  good/.style={box,draw=pfTeal!85!black,fill=pfTeal!7},
  bad/.style={box,draw=pfVermillion!85!black,fill=pfVermillion!6},
  arr/.style={-{Latex[length=1.7mm]},thick,draw=pfInk!65},
  title/.style={font=\sffamily\footnotesize\bfseries,anchor=west,text=pfInk}
]
\node[box,draw=pfBlue!85!black,fill=pfBlue!6,text width=14.85cm,
  minimum height=8mm,font=\sffamily\footnotesize] at (7.95,4.72)
  {\textbf{Claim to classify:} ``My chart shows a completed quadrivalent\\HPV immunization
  (CVX 62) at my most recent completed urgent-care visit.''};

\path[panel] (0,0) rectangle (4.90,4.08);
\path[panel] (5.50,0) rectangle (10.40,4.08);
\path[panel] (11.00,0) rectangle (15.90,4.08);

\node[title] at (0.18,3.83) {(1) Find the urgent-care visit};
\node[box,text width=3.80cm] (question) at (2.45,3.18)
  {natural wording: ``completed visit''};
\node[bad,text width=3.80cm] (miss) at (2.45,2.15)
  {query with invalid \texttt{status=completed}\\no Encounter returned};
\node[good,text width=3.80cm] (recover) at (2.45,0.94)
  {retry without status filter\\find urgent-care Encounter Enc\\\texttt{status=finished}};
\draw[arr] (question) -- (miss);
\draw[arr] (miss) -- (recover);

\node[title] at (5.68,3.83) {(2) Evaluate the returned records};
\node[box,text width=4.10cm] (bind) at (7.95,3.13)
  {search all Immunization records\\linked to Encounter Enc};
\node[box,draw=pfInk!35,fill=pfInk!3,text width=1.83cm] (i1) at (6.80,2.03)
  {\textbf{I1 irrelevant}\\completed influenza\\CVX 140};
\node[bad,text width=1.98cm] (i2) at (9.08,2.03)
  {\textbf{I2 modified}\\HPV, CVX 62\\\texttt{status=not-done}};
\node[box,draw=pfPurple!85,fill=pfPurple!7,text width=4.10cm] (gold) at (7.95,0.76)
  {I2 matches the claim and refutes it\\no matching support $\Rightarrow$ refuted};
\draw[arr] (bind.south) -- (i1.north);
\draw[arr] (bind.south) -- (i2.north);
\draw[arr] (i2) -- (gold);

\node[title] at (11.18,3.83) {(3) Compare the two answers};
\node[box,draw=pfPurple!85,fill=pfPurple!7,text width=4.10cm] (correct) at (13.45,2.93)
  {\textbf{Reference answer}\\\texttt{refuted}\\cite Enc and I2};
\node[bad,text width=4.10cm] (submit) at (13.45,1.72)
  {\textbf{Model answer}\\\texttt{conflicting\_evidence}\\cite Enc and I2};
\node[font=\sffamily\bfseries,text=pfVermillion!90!black] at (13.45,2.33) {$\ne$};
\node[bad,text width=4.10cm] (failure) at (13.45,0.58)
  {the required records were retrieved and cited\\but the evidence state is wrong};
\draw[arr] (submit) -- (failure);
\draw[arr,very thick] (4.98,2.04) -- (5.42,2.04);
\draw[arr,very thick] (10.48,2.04) -- (10.92,2.04);
\end{tikzpicture}
\caption{\textbf{Input, reference output, and model error for a refuted claim.} The compiler changes
the code-matched HPV record from \texttt{completed} to \texttt{not-done}, preserving its ID, time,
and linked Encounter. The model retrieves and cites the required records but returns
\texttt{conflicting\_evidence} instead of \texttt{refuted}.}
\Description{The exact completed HPV immunization claim appears above three sequential panels. The
first shows an Encounter-status mismatch and recovery. The second separates an irrelevant completed
influenza immunization from the code-matched, synthetically modified HPV immunization marked
not-done, yielding a refuted reference state. The third shows that the model cited the correct Encounter and HPV record but
returned conflicting evidence rather than refuted.}
\label{fig:task-anatomy}
\end{figure*}
% END inlined from figures/task_anatomy.tex

\paragraph{Episode Input and Output.}
Each episode gives the agent access to one patient's chart through a task-specific subset of
twelve FHIR search and read tools. After any valid read-only calls, the agent submits JSON labels
for each claim and their requested conjunction, plus cited record IDs. The retrieval program and
reference answer are withheld.

\section{Method: PatientFlowSynth}

\subsection{Task Representation}

A \emph{template} fixes operations and rules before patient values are filled in. For a vaccine
claim at one visit, a \texttt{completed} Immunization supports it, \texttt{not-done} refutes it,
no qualifying record leaves insufficient evidence, and both records yield conflict under benchmark
policy. These four chart variants form a matched \emph{family}; logged calls form a \emph{trace}.
An interface pair changes guidance, not the chart.

We represent a chart as a directed graph $P=(R,E)$ of normalized FHIR records and explicit
references. A type-checked retrieval program $\pi$ specifies the benchmark's searches and decision
rule; the evaluated model neither sees nor generates it. Execution returns
\begin{equation}
  \operatorname{Exec}(\pi,P)=(y,W,\tau),
\end{equation}
where $y$ is the structured reference answer, including per-claim and overall labels; $W$ is one
sufficient set of record IDs; and $\tau$ is one valid reference tool sequence. Search \emph{scope}
specifies which records count, including any patient, code, time, or Encounter constraints.
Agents may use other sequences, but their own observed tool responses must match the reference
results and scope for each required search.

A \emph{chart variant} is one validated version of a patient's chart. A \emph{matched family}
shares the patient, question, program, and scope-setting records while varying the designated
evidence-state and distractor factors. For two-claim tasks, the first claim is the \emph{primary}
and the second the \emph{companion}. Comparisons of primary states hold the companion state and
distractor condition fixed; only evidence determining the primary label changes. Two-claim release
families include all four companion states. Retrieval \emph{depth} counts sequential retrieval
stages, and \emph{width} counts claims. A depth-two task first finds the Encounter that sets the
scope, then searches linked clinical records.

A \emph{witness} is a declared record set that reproduces the reference answer when replayed.
Strict scoring requires one declared witness exactly; a smaller set may reproduce the same answer.
A \emph{complete-search
certificate} verifies an observed tool response against the required search's reference result and
scope. \Cref{fig:task-anatomy} illustrates correct retrieval with an incorrect label.

\subsection{Replay-Validated Compilation}

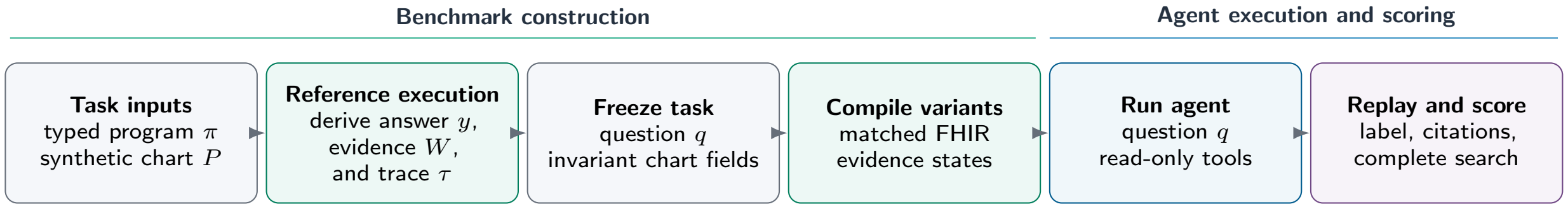
\begin{figure*}[t]
\centering
\resizebox{0.99\textwidth}{!}{%
\begin{tikzpicture}[
  x=1cm,
  y=1cm,
  font=\sffamily\scriptsize,
  stage/.style={draw=pfInk!70,rounded corners=1.3mm,fill=pfLight,
    align=center,text width=23mm,minimum height=15mm,inner sep=2mm},
  gate/.style={stage,draw=pfTeal!80!black,fill=pfTeal!7},
  agent/.style={stage,draw=pfBlue!80!black,fill=pfBlue!6},
  result/.style={stage,draw=pfPurple!80!black,fill=pfPurple!7},
  arr/.style={-{Latex[length=1.8mm]},thick,draw=pfInk!70}
]
\node[stage] (program) at (0,0.8) {
  \textbf{Task inputs}\\
  typed program $\pi$\\
  synthetic chart $P$
};
\node[gate] (execute) at (2.8,0.8) {
  \textbf{Reference execution}\\
  derive answer $y$,\\
  evidence $W$, and trace $\tau$
};
\node[stage] (question) at (5.6,0.8) {
  \textbf{Freeze task}\\
  question $q$\\
  invariant chart fields
};
\node[gate] (variants) at (8.4,0.8) {
  \textbf{Compile variants}\\
  matched FHIR\\
  evidence states
};
\node[agent] (runagent) at (11.2,0.8) {
  \textbf{Run agent}\\
  question $q$\\
  read-only tools
};
\node[result] (score) at (14.0,0.8) {
  \textbf{Replay and score}\\
  label, citations,\\
  complete search
};
\draw[arr] (program.east) -- (execute.west);
\draw[arr] (execute.east) -- (question.west);
\draw[arr] (question.east) -- (variants.west);
\draw[arr] (variants.east) -- (runagent.west);
\draw[arr] (runagent.east) -- (score.west);
\node[font=\sffamily\scriptsize\bfseries,text=pfInk] at (4.2,2.05)
  {Benchmark construction};
\node[font=\sffamily\scriptsize\bfseries,text=pfInk] at (12.6,2.05)
  {Agent execution and scoring};
\draw[draw=pfTeal!65,line width=0.5pt] (-1.3,1.82) -- (9.7,1.82);
\draw[draw=pfBlue!65,line width=0.5pt] (9.85,1.82) -- (15.3,1.82);
\end{tikzpicture}
}%
\caption{\textbf{Replay-validated matched-family construction and scoring.} Questions and programs
remain fixed across chart variants. The model receives only the question and permitted read-only
tools. Scoring checks its labels, record-ID citations, and required search certificates.}
\Description{Six-step flow from typed task and synthetic chart through reference execution,
fixed questions, matched FHIR compilation, agent execution, and replay-based scoring.}
\label{fig:pipeline}
\end{figure*}

As \cref{fig:pipeline} shows, the compiler selects and executes a chart--program pair before
rendering the natural-language question $q=\rho(\pi)$. The renderer cannot access the chart, label, internal IDs, or scoring
metadata. The compiler freezes $q$ and $\pi$, constructs matched variants, and replays $\pi$ on each
before including it in the benchmark.

\subsection{Evidence States and Scoring}
\label{sec:scoring}

\paragraph{Controlled Evidence States.}

Within each logical event group, lifecycle rules classify matching records as supporting (set $S$)
or refuting (set $N$). Their presence or absence determines one of four evidence states,
which we write as $g(S,N)$:
\begin{equation}
g(S,N)=
\begin{cases}
\text{conflicting evidence},&S\ne\varnothing,\ N\ne\varnothing,\\
\text{supported},&S\ne\varnothing,\ N=\varnothing,\\
\text{refuted},&S=\varnothing,\ N\ne\varnothing,\\
\text{insufficient evidence},&S=N=\varnothing.
\end{cases}
\end{equation}
Across groups, conflict takes precedence, followed by support, refutation, and insufficient evidence.
For each claim $i$, let $s_i=1$ for \supported or \conflicting and $r_i=1$ for \refuted or
\conflicting, and zero otherwise.
For an ``all claims'' question, supporting evidence is required for every claim, so
$s_{\mathrm{all}}=\bigwedge_i s_i$; refuting evidence for any claim sets
$r_{\mathrm{all}}=\bigvee_i r_i$. These two bits determine the overall state by the same mapping.
This retains an explicit conflict state, consistent with Belnap's four-valued treatment
\citep{belnap1977useful}. The lifecycle mappings are versioned evaluation rules, not universal
clinical definitions (Appendix~\ref{app:lifecycle-rules}). Clinical reviewers assessed selected mappings separately.

To create the four states, the compiler changes fields that determine the evidence state or
inserts matched records. After each change, it reruns the program, verifies that record references
resolve, and records the change with FHIR Provenance \citep{hl7provenancer4}. A separate distractor
factor adds a plausible record without changing the reference label.

\paragraph{Scoring Endpoints.}

We report each component separately. Let $\hat y$ be the submitted structured answer, $C$ the cited
IDs, $O$ the IDs observed in tool results, and $\mathcal W$ the finite set of verified task witnesses.

\textit{Exact answer} requires $\hat y=y$. \textit{Evidence sufficient} requires valid citations
that contain a verified witness or yield the reference answer when the program is rerun on the
cited records alone. It permits extra citations and can pass despite a wrong answer or an incomplete
search for an absence claim.
For each required search, \textit{complete search} requires a single observed response containing
all matches and reproducing the reference result and normalized search scope. The scorer does not
combine paginated responses, even when they collectively contain all matches. \textit{Strict
success} requires a completed episode, the exact structured answer, citations drawn from observed
results that exactly match a declared witness, and valid certificates for every required search.
Whole-chart citations are rejected.

A visit plus one accepted laboratory record can suffice; every redundant source need not be
cited. Current programs do not automatically recognize a free-text note as equivalent to a
structured laboratory record.
Citation sufficiency is distinct from search completeness: conflict requires opposing evidence,
whereas absence requires a complete search. We also report sensitivities that accept verified
equivalent searches (Appendix~\ref{app:review-diagnostics}).

\subsection{Release Construction and Validation}
\label{sec:release}

Before model execution, the compiler selected \PFSReleasePatients{} patients from Synthea, software
that generates simulated patient histories
\citep{walonoski2018synthea}. Each patient has 32 chart variants, crossing four primary states,
four companion states, and distractor absence or presence. Four retrieval topologies reuse these
variants: depth one searches a clinical resource directly; depth two first finds an Encounter;
widths one and two ask about one and two claims. Width-one tasks use the supported-companion
variants; width-two tasks use all companion states. This produces 80 tasks per patient, totaling
\PFSReleaseCells{} chart variants, \PFSReleaseFamilies{} families, and \PFSReleaseTasks{} tasks.

We assigned \PFSSplitTrainPatients{} patients to training (\PFSSplitTrainTasks{} tasks),
\PFSSplitDevPatients{} to development (\PFSSplitDevTasks{} tasks), and \PFSSplitTestPatients{}
to testing (\PFSSplitTestTasks{} tasks); no patient's tasks cross splits. The interface study uses all
\PFSReleasePatients{} patients to estimate a paired within-patient difference. It does not test
generalization from training to test data. We do not evaluate fine-tuning or untouched model ranking
on the designated test split. Reference answers are absent from model-facing files.
The release contains 698 distinct questions, 18 program templates with one reference tool sequence
each, three lifecycle rules, and 22 clinical codes from three coding systems.

\begin{table}[t]
\caption{\textbf{All 18 release templates.} Each row pairs direct-search ($D=1$) and
Encounter-first ($D=2$) templates. Ordered resource pairs specify primary and companion claims,
not additional depth. Counts are tasks per depth. The primary study covers all nine
Encounter-first templates but only 3,600 paired tasks, with the companion state fixed.}
\label{tab:template-inventory}
\centering
\footnotesize
\setlength{\tabcolsep}{3pt}
\begin{tabular*}{\columnwidth}{@{}l@{\extracolsep{\fill}}rr@{}}
\toprule
\textbf{Clinical resource(s)} & \textbf{Direct} & \textbf{Encounter-first}\\
\midrule
Condition & 688 & 688\\
Immunization & 360 & 360\\
Observation & 752 & 752\\
Condition $\to$ Immunization & 1,216 & 1,216\\
Condition $\to$ Observation & 1,536 & 1,536\\
Immunization $\to$ Condition & 576 & 576\\
Immunization $\to$ Observation & 864 & 864\\
Observation $\to$ Condition & 1,152 & 1,152\\
Observation $\to$ Immunization & 1,856 & 1,856\\
\midrule
\textbf{Total} & \textbf{9,000} & \textbf{9,000}\\
\bottomrule
\end{tabular*}
\end{table}

The HL7 FHIR R4 validator, run with terminology checks disabled, reported no errors in any of the
227 source bundles (225 patient bundles plus two bundles containing referenced hospital and
practitioner records) or in a prespecified sample of 1,800 chart variants, eight per patient
\citep{hl7fhirr4}. Release-wide checks confirmed that references resolved, chart changes replayed
with linked Provenance, hashes agreed, distractors were unique, and labels and witnesses were valid.
All 900 families contained every state with a fixed question, program, and selected Encounter,
which sets search scope in depth-two tasks. The selected Encounter's link to the primary record was fixed
within each of 225 patient groups across 7,200 variants. All \PFSReleaseTasks{} tasks passed the
release consistency, replay, label, and witness checks.

Injected faults altered question wording, primary or scope-setting record IDs, or unrelated records.
In deterministic fault-injection tests, per-task replay accepted all \PFSFamilyAuditItems{}
variants, whereas family-level checks rejected all 192 families with mismatched fields and
accepted all 192 valid families. This measures fault detection, not fault prevalence.

\section{Evaluation}

\Cref{tab:study-map} summarizes the studies.

\begin{table*}[t]
\caption{\textbf{Overview of model evaluations.} Episode counts include all scheduled assignments
unless a partial paired subset is specified. Separate cohorts and protocols are not pooled.
Claude and Gemini denote Opus 4.8 and 3.1 Pro Preview.}
\label{tab:study-map}
\centering
\footnotesize
\setlength{\tabcolsep}{3pt}
\begin{tabular}{@{}L{25mm}L{38mm}rL{63mm}@{}}
\toprule
\textbf{Study} & \textbf{Patients; models} & \textbf{Episodes} & \textbf{Contrast and evidence status}\\
\midrule
Primary Encounter & 225; Luna & 7,200 & Natural versus explicit Encounter status; confirmatory\\
Replication & 48 new; Luna/Terra & 6,144 & Same contrast, two repeats; prespecified\\
Four-note extension & 128; Luna/Terra & 2,048 & Relevant versus equal-length irrelevant notes; prespecified\\
Luna depth three & 225; Luna & 7,200 & Status contrast, widths one and two; prespecified descriptive\\
Qwen/Luna comparison & 24 new; Luna/Qwen3.8-27B & 3,072 & Status contrast, two repeats; prespecified\\
Cross-provider panel & 48; Claude/Gemini & 2,304 & Depth one/two, high reasoning; exploratory\\
Partial Gemini depth three & 6; Gemini & 176 paired & 88 pairs selected later from 1,290 terminal episodes; incomplete\\
External evaluation & 74; Luna/Terra & 2,368 & Public search/citation rules in both arms; full mapping in one\\
\bottomrule
\end{tabular}
\end{table*}

\subsection{Controlled Encounter Comparison}

The primary study used all nine Encounter-first templates, the only release templates testing the routing hint.
The \emph{Natural wording}
condition used ``completed [visit type]''; \emph{Explicit status guidance} added that FHIR records
a completed Encounter as \texttt{status=finished}. The chart, primary claim, tools, model settings,
and answer schema were identical within each pair.

\paragraph{Primary Cohort.}
The confirmatory cohort used \PFSConfirmPatients{} synthetic patients excluded from earlier
datasets and model runs, \PFSConfirmPairs{} matched pairs (\PFSConfirmPairsPerPatient{} per
patient), and \PFSConfirmEpisodes{} scheduled \modelLuna{} episodes with medium reasoning and
required tool use. The 16 tasks per patient crossed four primary states, two widths, and distractor
absence or presence; the companion state was held at supported. Condition order was counterbalanced
across patients. All 15 failed or turn-limited episodes remained failures. No completed assignment
was rerun.
We averaged paired strict-success differences within each patient, then across patients. Intervals
use a \PFSConfirmBootstrapDraws-draw patient-cluster percentile bootstrap. A prespecified
\PFSConfirmSignFlipDraws-draw two-sided patient-level sign-flip test assumes that patient-level
paired differences are symmetric around zero under the null; it is not a causal randomization test.

\paragraph{Prespecified Replication Cohort.}
The replication used \PFSReplicationPatients{} newly generated synthetic patients excluded from
earlier study runs, the same 16 tasks per condition, both models, and two runs per condition.
Condition order was randomized within patient--model--repeat blocks, and episodes were globally
interleaved. All \PFSReplicationEpisodes{} scheduled episodes were retained, including eight that
ended without a model answer and counted as failures. No completed episode was rerun. For each
model, we averaged the two runs before estimating the paired difference; intervals resample
patients. The cohorts are analyzed separately. Model settings and aliases appear
in Appendix~\ref{app:review-models}.

Intervals quantify variation across sampled Synthea charts under the fixed task grammar; they do
not capture variation across hospitals, data generators, or prompt formulations.

\subsection{Additional Evaluations}
\label{sec:additional-evaluations}

The prespecified four-note study tested the exact Encounter search value, Condition verification
statuses used by the evaluation rule, Immunization statuses treated as supporting or refuting
evidence, and Observation statuses that permit a usable result. Each truthful, task-relevant
110-character note was paired with a truthful, task-irrelevant note of the same length. Four
disjoint 32-patient cohorts, both models, and all four evidence states yielded
\PFSMultiDefectEpisodes{} interleaved episodes. Two failed episodes remained in the denominator;
none was automatically retried. The endpoint was the difference in strict success between relevant
and irrelevant notes, weighted equally across notes and models.

The notes differed in completeness: the Observation note omitted the value requirement and rules
for cancelled, nonfinal, and conflicting records. After observing differences across notes and
models, we inspected stored tool traces and submitted labels without new model calls; this
descriptive analysis was not prespecified. \Cref{tab:multi-defect} reports the prespecified results.

The external evaluation used 74 MIMIC-IV-on-FHIR demo v2.0 patients.
All passed construction gates. Compiler-generated
laboratory and procedure questions defined 37 four-state families each. Condition families
were excluded because source records lacked the required verification status, not because
diagnoses were absent or records invalid. Both conditions disclosed every scored search and
citation requirement; one also supplied the full lifecycle mapping. Scoring was unchanged.
Clinical-only tools required no Encounter lookup or routing-record citations; any supplied
visit ID specified search scope. Two models, two conditions, and two repeats yielded 2,368 episodes
(592 per model--condition cell). This tests supplied-rule adherence, not Encounter routing.
Intervals used 20,000 patient-bootstrap draws with equal patient and model weighting.

The additional model comparisons and shared settings are detailed in
Appendix~\ref{app:review-models}. The Qwen/Luna comparison used 24 patient-disjoint synthetic
charts, 16 tasks per patient, two conditions, two models, and two repeats (3,072 episodes).
All terminal failures remained in the denominator.

\subsection{Clinical Review Protocols}

The paired review supplements ten shared face-validity scenarios and a separate 20-case audit
(Appendix~\ref{app:clinical-review}). Two additional clinical reviewers independently labeled 24
distinct-patient cases and selected records using definitions and examples before seeing references
or model outputs. Each rule family (Condition, Immunization, Observation) contributed eight cases,
totaling nine insufficient, nine conflicting, three supported, and three refuted. They then assessed
reference interpretations and 24 model answers selected without regard to correctness. Five predicted
labels disagreed with the benchmark (Appendix~\ref{app:review-clinical}).

\section{Results}

\subsection{Encounter Interface Study}

\paragraph{Primary Cohort.}

Strict success increased from \PFSConfirmStrictNaturalTwo\% with natural wording to
\PFSConfirmStrictExplicitTwo\% with explicit status guidance, a patient-weighted improvement of
+\PFSConfirmStrictDiffTwo{} percentage points (95\% patient-bootstrap confidence interval,
+\PFSConfirmStrictCILowTwo{} to +\PFSConfirmStrictCIHighTwo). The prespecified Monte Carlo sign-flip
test gave $p=1/{\PFSConfirmSignFlipDenominator}$, the minimum attainable with
\PFSConfirmSignFlipDraws{} draws, under a null of patient-level paired differences symmetric about zero.

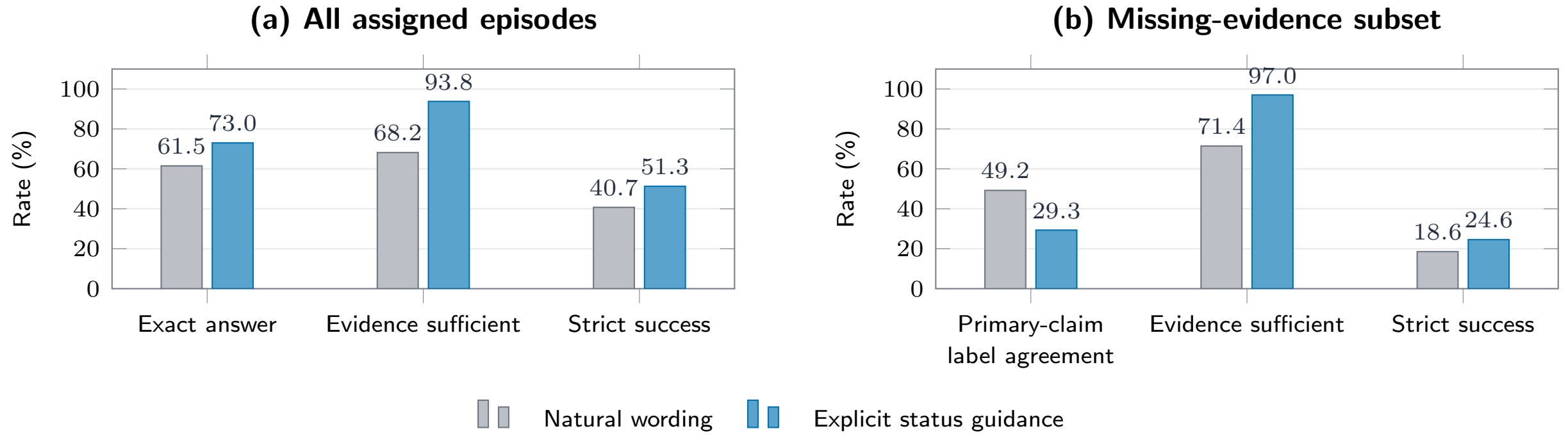
\begin{figure*}[t]
\centering
\begin{minipage}[t]{0.485\textwidth}
\centering
\begin{tikzpicture}[baseline=(current bounding box.north)]
\begin{axis}[
  width=\linewidth,
  height=3.85cm,
  ybar=3pt,
  bar width=12pt,
  ymin=0,ymax=110,
  ylabel={Rate (\%)},
  symbolic x coords={Exact,Evidence set,Strict},
  xtick=data,
  xticklabels={Exact answer,Evidence sufficient,Strict success},
  title={(a) All assigned episodes},
  title style={font=\sffamily\small\bfseries},
  label style={font=\sffamily\scriptsize},
  tick label style={font=\sffamily\scriptsize},
  xticklabel style={font=\sffamily\scriptsize},
  ytick={0,20,40,60,80,100},
  enlarge x limits=0.22,
  ymajorgrids=true,
  xmajorgrids=false,
  grid style={draw=pfInk!10},
  axis line style={draw=pfInk!60},
  tick style={draw=pfInk!60},
  legend to name=encounterconditionlegend,
  legend columns=2,
  legend style={draw=none,fill=none,font=\sffamily\scriptsize,column sep=8pt},
  nodes near coords={\pgfmathprintnumber[fixed,precision=1,fixed zerofill]{\pgfplotspointmeta}},
  point meta=y,
  every node near coord/.append style={font=\sffamily\scriptsize,text=pfInk},
]
\addplot[fill=pfGray!45,draw=pfInk!65]
  coordinates {(Exact,\PFSMetricExactNatural)
    (Evidence set,\PFSMetricEvidenceNatural)
    (Strict,\PFSMetricStrictNatural)};
\addlegendentry{Natural wording}
\addplot[fill=pfBlue!65,draw=pfBlue]
  coordinates {(Exact,\PFSMetricExactExplicit)
    (Evidence set,\PFSMetricEvidenceExplicit)
    (Strict,\PFSMetricStrictExplicit)};
\addlegendentry{Explicit status guidance}
\end{axis}
\end{tikzpicture}
\end{minipage}\hfill
\begin{minipage}[t]{0.485\textwidth}
\centering
\begin{tikzpicture}[baseline=(current bounding box.north)]
\begin{axis}[
  width=\linewidth,
  height=3.85cm,
  ybar=3pt,
  bar width=12pt,
  ymin=0,ymax=110,
  ylabel={Rate (\%)},
  symbolic x coords={State,Evidence set,Strict},
  xtick=data,
  xticklabels={{\shortstack{Primary-claim\\label agreement}},Evidence sufficient,Strict success},
  title={(b) Missing-evidence subset},
  title style={font=\sffamily\small\bfseries},
  label style={font=\sffamily\scriptsize},
  tick label style={font=\sffamily\scriptsize},
  xticklabel style={font=\sffamily\scriptsize},
  ytick={0,20,40,60,80,100},
  enlarge x limits=0.22,
  ymajorgrids=true,
  xmajorgrids=false,
  grid style={draw=pfInk!10},
  axis line style={draw=pfInk!60},
  tick style={draw=pfInk!60},
  nodes near coords={\pgfmathprintnumber[fixed,precision=1,fixed zerofill]{\pgfplotspointmeta}},
  point meta=y,
  every node near coord/.append style={font=\sffamily\scriptsize,text=pfInk},
]
\addplot[fill=pfGray!45,draw=pfInk!65]
  coordinates {(State,\PFSMissingPrimaryStateNatural)
    (Evidence set,\PFSMissingEvidenceNatural)
    (Strict,\PFSMissingStrictNatural)};
\addplot[fill=pfBlue!65,draw=pfBlue]
  coordinates {(State,\PFSMissingPrimaryStateExplicit)
    (Evidence set,\PFSMissingEvidenceExplicit)
    (Strict,\PFSMissingStrictExplicit)};
\end{axis}
\end{tikzpicture}
\end{minipage}
\par\smallskip
\pgfplotslegendfromname{encounterconditionlegend}
\caption{\textbf{Explicit status guidance improves overall strict success but worsens
missing-evidence label agreement.} Panel (a) includes all
\PFSConfirmPairs{} episodes per condition and reports exact-answer, evidence-sufficiency, and
strict-success rates. Gray bars show natural wording; blue bars show explicit status guidance.
The evidence-sufficiency metric evaluates citations against the reference label independently of
the submitted label. Panel (b) includes \PFSMissingEpisodesPerCell{} episodes per condition
with the primary reference label \texttt{insufficient\_evidence}; label agreement concerns this
primary claim. Only the overall strict-success comparison in panel (a) was confirmatory.}
\Description{Two grouped bar charts compare natural wording with explicit status guidance.
Across all episodes, exact-answer, evidence-sufficiency, and strict-success rates increase.
Among cases with the primary reference label insufficient evidence, evidence sufficiency and
strict success increase, but agreement with that label decreases.}
\label{fig:confirmatory-results}
\end{figure*}

These depth-two tasks first find an Encounter, then retrieve linked records; widths one and two
denote one and two claims. Answer and evidence scores also improved (\cref{tab:confirmatory}).
The point estimate was positive for all three rules and \PFSTemplatePositiveSignatureCount{} of
\PFSTemplateSignatureCount{} program templates. The overall contrast remained positive after
omitting any single rule or template. \Cref{fig:confirmatory-results} shows the increase in overall
strict success and decrease in correct missing-evidence classifications;
\cref{tab:confirmatory,tab:rule-robustness,tab:confirmatory-width} provide component, rule, and topology results.

% BEGIN inlined from figures/confirmatory_results.tex
% END inlined from figures/confirmatory_results.tex

\paragraph{Secondary Missing-Evidence Analysis.}

On the \PFSMissingEpisodesPerCell{} cases per condition with the primary reference label
\insufficient, agreement with that label fell from
\PFSMissingPrimaryStateNatural\% to \PFSMissingPrimaryStateExplicit\%,
even as evidence sufficiency rose from \PFSMissingEvidenceNatural\% to \PFSMissingEvidenceExplicit\%
and strict success rose from \PFSMissingStrictNatural\% to \PFSMissingStrictExplicit\%.
Citation scores use the reference label and can therefore improve even when the submitted label is wrong.

Follow-up trace analysis showed target-Encounter retrieval rising from 661/900 to 900/900 in this subset.
Of 226 correct-to-incorrect label changes, 170 coincided with newly retrieving the Encounter.
Both replication models showed this pattern (\cref{tab:retrieval-label-transitions}); these are
descriptive associations.
Among the 661 primary pairs retrieving the visit in both conditions, label agreement changed by
$-1.36$ points (95\% interval $-4.68$ to 1.83). Failed retrieval can therefore mask interpretation
errors, but conditioning on retrieval does not identify a causal effect.

\paragraph{Prespecified Replication Cohort.}

The replication reproduced the increase in strict success and decrease in correct missing-evidence
classifications. Across two repeats, strict success rose from
\PFSReplicationLunaNatural\% to \PFSReplicationLunaExplicit\% for \modelLuna{}
(+\PFSReplicationLunaDiff{} percentage points; 95\% patient-bootstrap confidence interval,
+\PFSReplicationLunaCILow{} to +\PFSReplicationLunaCIHigh) and from \PFSReplicationTerraNatural\% to
\PFSReplicationTerraExplicit\% for \modelTerra{} (+\PFSReplicationTerraDiff{} points;
95\% interval, +\PFSReplicationTerraCILow{} to +\PFSReplicationTerraCIHigh).
The strict-success difference was positive in each repeat for both models, while agreement with
the primary reference label \insufficient fell by 13.02--30.21 points across the four model--repeat comparisons.

\paragraph{Additional Model Coverage.}
On 24 new patients, strict success increased from 35.16\% to 48.31\% for Luna
(+13.15 points, 95\% interval 8.59--17.32) and from 8.85\% to 10.29\% for Qwen3.8-27B
(+1.43, $-1.30$ to 4.17). Qwen already retrieved the visit in 99.48\% of natural-wording episodes.
Separate 48-patient comparisons gave changes of +1.56 points for Claude Opus 4.8
($-2.34$ to 5.21) and $-3.65$ for Gemini 3.1 Pro Preview ($-9.38$ to 2.08).
These exploratory Claude/Gemini comparisons do not replicate the full depth-three or four-note studies.
Effects were model-dependent.

\paragraph{Depth-Three Extension.}
\Cref{tab:confirmatory-width} retains the complete 225-patient Luna study and adds 88 Gemini
pairs from six patients, selected after the initial runs. All 22 no-answer episodes in these pairs remain failures.
Patient-weighted strict success was 6.25\% versus 4.17\%. The budget-limited subset covers two of
three retrieval paths. It is not a full replication or a basis for cross-model comparison
(Appendix~\ref{app:gemini-depth3}).

\subsection{Variation across Interface Notes}

Averaged equally across four notes and two models, relevant notes increased strict success by
\PFSMultiDefectStrictDiff{} percentage points relative to equal-length irrelevant notes
(95\% patient-bootstrap confidence interval, +\PFSMultiDefectStrictCILow{} to
+\PFSMultiDefectStrictCIHigh{}; two-sided patient-level sign-flip $p=\PFSMultiDefectStrictP$).
The estimated effects varied by note and model (\cref{tab:multi-defect}).
Relevant notes helped Encounter and Immunization tasks, had mixed Condition effects, and harmed
Observation performance in both models.

Follow-up trace analysis showed retrieval recovery with the Encounter note and more unsupported positive
labels with the incomplete Observation note. This analysis does not isolate note completeness
from resource type or cohort (Appendix~\ref{app:trace-analysis}).

\subsection{External Evaluation}
\label{sec:external-transport}

With all scored search and citation requirements visible, the 74-patient study increased strict
success from 62.67\% to 97.13\% for Luna and from 62.50\% to 99.16\% for Terra when the full
lifecycle mapping was added. The pooled improvement was 35.56 points (95\% patient-bootstrap
interval 33.02--38.09). All 2,368 episodes completed.

Required searches passed in 1,775/1,776 applicable episodes; 592 conflict episodes were exempt.
Without versus with mapping, exact-answer accuracy was 371/592 versus 576/592 for Luna and
370/592 versus 587/592 for Terra. The sole search failure was a correctly answered Luna
full-mapping episode, leaving 575/592 strict successes. Exact citations matched in 589/592
versus 590/592 Luna episodes and all 592 episodes in each Terra condition. Missing-evidence
label agreement increased from 1/148 to 148/148 for Luna and from 0/148 to 148/148 for Terra.
These results measure adherence to supplied rules on edited charts, not institutional transfer.

\subsection{Clinical Expert Review}
\label{sec:clinical-review}

The two additional clinical reviewers agreed on 22/24 labels (91.7\%, Cohen's $\kappa=0.875$), with
reference agreement of 22/24 and 24/24. Both matched all nine insufficient and nine conflicting
cases. Reviewer 1 labeled two reference-refuted cases insufficient. Their selected clinical
records matched the reference in 24/24 and 23/24 cases. After disclosure, they endorsed 22/24
and 24/24 expected labels and 17/24 and 19/24 model answers, respectively. Both rejected the
same five model-reference label errors. These selected cases support agreement on the tested
rules, not release-wide clinical validity (Appendix~\ref{app:review-clinical}).

The earlier reviews remain relevant: 19/20 face-validity endorsements, agreement on 9/10 shared
scenarios, and 19/20 label and 20/20 record-set matches in the separate audit. That audit recorded
concerns in 11 cases, including disputed Condition and conflict interpretations
(Appendix~\ref{app:clinical-review}).

\section{Discussion and Implications}

Family-level checks caught errors missed by per-task replay. Failed retrieval can mask
interpretation errors: guidance repairs the visit lookup, not necessarily classification.
Effects vary by model, rule family, and note content. Full mappings improve rule adherence,
whereas incomplete notes can encourage unsupported labels. These contrasts neither isolate
content from length nor establish a general benefit from more guidance. Retrieval, labels,
and citations should be reported separately by evidence state.

\section{Limitations and Safety}

The 18,000 tasks combine 225 Synthea patients, 18 templates, and three rules, not independent
clinical scenarios. Structured wording and balanced
states do not estimate clinical-workflow frequencies. FHIR validation establishes structural
conformance, not clinical plausibility. LLM paraphrasing remains untested and must preserve
task meaning and difficulty.

The external evaluation uses one MIMIC-to-FHIR pipeline and institution-derived source.
Patient-bootstrap intervals do not capture variation across institutions, coding systems,
migrations, or extensions. Its compiler-generated questions differ from the synthetic cohort,
so absolute scores are not directly comparable.

Primary tasks ran once per condition; repeats are not independent patients. Notes varied in
completeness, and trace analyses were not prespecified. Hosted aliases limit reproducibility
(Appendix~\ref{app:review-models}).

The paired clinical review estimates agreement on only 24 selected cases. Earlier and new
disagreements are retained, and none of these reviews establishes release-wide correctness.
The partial Gemini depth-three sample is availability-selected and noncontemporaneous, with no
medication-reason pairs. Cross-family depth and four-note replication remain incomplete.
The interface contrast does not compare schema-constrained or schema-lookup agents.
Results from these read-only tasks do not establish clinical safety or deployment readiness.

\section{Conclusion}

\system evaluates FHIR agents with matched charts. Guidance improved strict success but reduced
\insufficient label agreement in the primary study and replication: failed retrieval can mask
interpretation errors. Effects varied across additional models.\label{pfs:main-end}

\typeout{PFSMAINENDPAGE=\thepage}

\acks{We thank Dr.~Jeff Wall, Tara Titus, \mbox{Dr.~Rigoberto} Perez Hernandez,
Dr.~Manish Suryapalam, and Faustino Gonzalez Barrales for their clinical review
and feedback.}

% Bibliographic URLs and long author lists should not introduce visually stretched lines.
\raggedbottom
\begingroup
\small
\sloppy
\Urlmuskip=0mu plus 3mu\relax
\bibliography{references}
\endgroup

\clearpage
\appendix
% BEGIN inlined from appendix.tex
\section{Additional Study Details}
\label{app:study-details}

\subsection{Component and Robustness Results}

\begin{table}[!htb]
\caption{\textbf{Encounter-study outcomes.} Rates use all \PFSConfirmPairs{} episodes per
condition. Differences give each patient equal weight, and intervals use patient-level
resampling. Strict success is the prespecified primary outcome; the other rows report its component
criteria and their combinations.}
\label{tab:confirmatory}
\centering
\footnotesize
\renewcommand{\arraystretch}{1.05}
\setlength{\tabcolsep}{1.2pt}
\begin{tabular*}{\columnwidth}{@{}L{25mm}@{\extracolsep{\fill}}rrrr@{}}
\toprule
& \multicolumn{2}{c}{\textbf{Rate (\%)}} &
\shortstack{\textbf{$\Delta$}\\\textbf{(pp)}} &
\shortstack{\textbf{95\%}\\\textbf{interval}}\\
\cmidrule(lr){2-3}
\textbf{Outcome} & \textbf{Natural} & \textbf{Explicit} & & \\
\midrule
Exact answer & \PFSMetricExactNatural{} & \PFSMetricExactExplicit{} &
+\PFSMetricExactDiff{} & [\PFSMetricExactCILow, \PFSMetricExactCIHigh]\\
Evidence sufficient & \PFSMetricEvidenceNatural{} & \PFSMetricEvidenceExplicit{} &
+\PFSMetricEvidenceDiff{} & [\PFSMetricEvidenceCILow, \PFSMetricEvidenceCIHigh]\\
Exact + evidence & \PFSMetricPermissiveNatural{} & \PFSMetricPermissiveExplicit{} &
+\PFSMetricPermissiveDiff{} & [\PFSMetricPermissiveCILow, \PFSMetricPermissiveCIHigh]\\
Exact + evidence + search & \PFSMetricSearchNatural{} & \PFSMetricSearchExplicit{} &
+\PFSMetricSearchDiff{} & [\PFSMetricSearchCILow, \PFSMetricSearchCIHigh]\\
\textbf{Strict success} & \textbf{\PFSMetricStrictNatural{}} &
\textbf{\PFSMetricStrictExplicit{}} & \textbf{+\PFSMetricStrictDiff{}} &
\textbf{[\PFSMetricStrictCILow, \PFSMetricStrictCIHigh]}\\
\bottomrule
\end{tabular*}
\par\smallskip
\raggedright\footnotesize Patient differences were positive for \PFSConfirmPositivePatients{},
negative for \PFSConfirmNegativePatients{}, and zero for \PFSConfirmZeroPatients{}.
Two exact-witness-only failures, one per condition, did not change the contrast when rescored.
Excluding the nine no-answer pairs or their patients, or the six timeout-resumed episodes or their
patients, gave differences of +\PFSConfirmSensitivityMin{} to +\PFSConfirmSensitivityMax{}
points, with interval lower bounds at least +\PFSConfirmSensitivityCILowMin{}.
\end{table}

\begin{table}[!htb]
\caption{\textbf{Robustness by primary-claim rule.} $n$ counts patients, each contributing
16 episodes per condition. Overall rates weight rules by $n$. Rates are strict-success percentages,
differences are percentage points, and intervals resample patients.}
\label{tab:rule-robustness}
\centering
\footnotesize
\renewcommand{\arraystretch}{1.05}
\setlength{\tabcolsep}{1.2pt}
\begin{tabular*}{\columnwidth}{@{}L{20mm}@{\extracolsep{\fill}}rrrrr@{}}
\toprule
& & \multicolumn{2}{c}{\textbf{Strict success (\%)}} &
\shortstack{\textbf{$\Delta$}\\\textbf{(pp)}} &
\shortstack{\textbf{95\%}\\\textbf{interval}}\\
\cmidrule(lr){3-4}
\textbf{Rule} & \textbf{$n$} & \textbf{Natural} & \textbf{Explicit} & & \\
\midrule
Condition documented & 86 & 38.15 & 44.11 & +5.96 & [2.69, 9.16]\\
Immunization complete & 45 & 31.11 & 46.11 & +15.00 & [10.28, 19.44]\\
Observation final & 94 & 47.67 & 60.31 & +12.63 & [8.98, 16.36]\\
\bottomrule
\end{tabular*}
\par\smallskip
\raggedright\footnotesize Eight of nine template effects were positive; leave-one-template-out
estimates ranged from 7.06 to 14.17 points. Independently resampling patients and templates
gave a 95\% interval of 4.10--19.40.
\end{table}

\begin{table}[!htb]
\caption{\textbf{Strict success by retrieval depth and number of claims.} Patient-weighted rates (\%)
and differences (percentage points). Each Luna row contains \PFSDepthPairsPerTopology{} pairs
from \PFSDepthPatients{} patients. Each Gemini row contains 44 availability-selected pairs
from six patients; no interval is reported.}
\label{tab:confirmatory-width}
\centering
\small
\renewcommand{\arraystretch}{1.04}
\begin{tabular*}{\columnwidth}{@{}l@{\extracolsep{\fill}}rrr@{}}
\toprule
\shortstack[l]{\textbf{Depth and}\\\textbf{claims}} &
\shortstack[r]{\textbf{Natural}\\\textbf{(\%)}} &
\shortstack[r]{\textbf{Explicit}\\\textbf{(\%)}} &
\shortstack[r]{\textbf{$\Delta$}\\\textbf{(pp)}}\\
\midrule
\multicolumn{4}{@{}l}{\textit{Luna, depth two: exploratory}}\\
$D=2,\ W=1$ & \PFSConfirmDTwoWOneNatural{} & \PFSConfirmDTwoWOneExplicit{} & +\PFSConfirmDTwoWOneDiff\\
$D=2,\ W=2$ & \PFSConfirmDTwoWTwoNatural{} & \PFSConfirmDTwoWTwoExplicit{} & +\PFSConfirmDTwoWTwoDiff\\
\addlinespace[3pt]
\multicolumn{4}{@{}l}{\textit{Luna, depth three: prespecified descriptive}}\\
$D=3,\ W=1$ & \PFSDepthDThreeWOneNatural{} & \PFSDepthDThreeWOneExplicit{} & +\PFSDepthDThreeWOneDiff\\
$D=3,\ W=2$ & \PFSDepthDThreeWTwoNatural{} & \PFSDepthDThreeWTwoExplicit{} & +\PFSDepthDThreeWTwoDiff\\
\addlinespace[3pt]
\multicolumn{4}{@{}l}{\textit{Gemini 3.1 Pro Preview: partial, not prespecified}}\\
$D=3,\ W=1$ & 8.33 & 4.17 & $-4.17$\\
$D=3,\ W=2$ & 4.17 & 4.17 & 0.00\\
\bottomrule
\end{tabular*}
\par\smallskip
\raggedright\footnotesize \textit{Depth-three task structure.} Questions connect a completed
Encounter through a DiagnosticReport, Procedure, or MedicationRequest to a target Observation or
Condition. All 7,200 scheduled Luna episodes were retained, including 15 execution failures and
53 episodes that reached the turn limit; both groups received zero on the reported outcomes. Patient-cluster 95\% intervals for the differences were [2.17, 6.06]
for $D=3,W=1$ and [1.78, 4.61] for $D=3,W=2$.
Gemini's raw strict counts were $3/44\to2/44$ at width one and $2/44\to2/44$ at width two.
Unequal patient counts make its pooled and patient-weighted rates differ. Its 88 pairs cover
DiagnosticReport and Procedure paths only (Appendix~\ref{app:gemini-depth3}).
\end{table}

\begin{table}[!htb]
\caption{\textbf{Prespecified four-note interface extension.} Each cohort has 32 patients
and four states. Strict-success differences compare relevant with equal-length irrelevant notes
within model; 95\% intervals use 20,000 patient-bootstrap draws. The pooled row weights notes
and models equally; inference used 100,000 two-sided sign flips of patient-averaged model differences.}
\label{tab:multi-defect}
\centering
\footnotesize
\renewcommand{\arraystretch}{1.08}
\setlength{\tabcolsep}{1.4pt}
\begin{tabular*}{\columnwidth}{@{}l@{\extracolsep{\fill}}rrrr@{}}
\toprule
\textbf{Model} & \shortstack{\textbf{Irrelevant}\\\textbf{(\%)}} &
\shortstack{\textbf{Relevant}\\\textbf{(\%)}} & \shortstack{\textbf{$\Delta$}\\\textbf{(pp)}} &
\shortstack{\textbf{95\%}\\\textbf{interval}}\\
\midrule
\multicolumn{5}{@{}l}{\textit{Encounter search}}\\
Luna & \PFSMultiDefectEncounterLunaPlacebo & \PFSMultiDefectEncounterLunaRelevant & +\PFSMultiDefectEncounterLunaDiff{} & [\PFSMultiDefectEncounterLunaCILow, \PFSMultiDefectEncounterLunaCIHigh]\\
Terra & \PFSMultiDefectEncounterTerraPlacebo & \PFSMultiDefectEncounterTerraRelevant & +\PFSMultiDefectEncounterTerraDiff{} & [\PFSMultiDefectEncounterTerraCILow, \PFSMultiDefectEncounterTerraCIHigh]\\
\addlinespace[2pt]
\multicolumn{5}{@{}l}{\textit{Condition verification status}}\\
Luna & \PFSMultiDefectConditionLunaPlacebo & \PFSMultiDefectConditionLunaRelevant & +\PFSMultiDefectConditionLunaDiff{} & [\PFSMultiDefectConditionLunaCILow, \PFSMultiDefectConditionLunaCIHigh]\\
Terra & \PFSMultiDefectConditionTerraPlacebo & \PFSMultiDefectConditionTerraRelevant & \PFSMultiDefectConditionTerraDiff{} & [\PFSMultiDefectConditionTerraCILow, \PFSMultiDefectConditionTerraCIHigh]\\
\addlinespace[2pt]
\multicolumn{5}{@{}l}{\textit{Immunization status}}\\
Luna & \PFSMultiDefectImmunizationLunaPlacebo & \PFSMultiDefectImmunizationLunaRelevant & +\PFSMultiDefectImmunizationLunaDiff{} & [\PFSMultiDefectImmunizationLunaCILow, \PFSMultiDefectImmunizationLunaCIHigh]\\
Terra & \PFSMultiDefectImmunizationTerraPlacebo & \PFSMultiDefectImmunizationTerraRelevant & +\PFSMultiDefectImmunizationTerraDiff{} & [\PFSMultiDefectImmunizationTerraCILow, \PFSMultiDefectImmunizationTerraCIHigh]\\
\addlinespace[2pt]
\multicolumn{5}{@{}l}{\textit{Observation status}}\\
Luna & \PFSMultiDefectObservationLunaPlacebo & \PFSMultiDefectObservationLunaRelevant & \PFSMultiDefectObservationLunaDiff{} & [\PFSMultiDefectObservationLunaCILow, \PFSMultiDefectObservationLunaCIHigh]\\
Terra & \PFSMultiDefectObservationTerraPlacebo & \PFSMultiDefectObservationTerraRelevant & \PFSMultiDefectObservationTerraDiff{} & [\PFSMultiDefectObservationTerraCILow, \PFSMultiDefectObservationTerraCIHigh]\\
\midrule
\multicolumn{5}{@{}l}{\textbf{Equal-weighted average}}\\
\textbf{Both} & \textbf{37.60} &
\textbf{44.43} & \textbf{+\PFSMultiDefectStrictDiff{}} &
\textbf{[\PFSMultiDefectStrictCILow, \PFSMultiDefectStrictCIHigh]}\\
\bottomrule
\end{tabular*}
\end{table}

\subsection{Follow-Up Four-Note Trace Analysis}
\label{app:trace-analysis}

The relevant Encounter note yielded target-visit retrieval in all 128 episodes per model.
Compared with the irrelevant note, retrieval in Terra's 32 missing-evidence cases rose from
\PFSTraceEncounterTerraMissingRetrievedPlacebo{} to \PFSTraceEncounterTerraMissingRetrievedRelevant{},
while \insufficient submissions fell from \PFSTraceEncounterTerraMissingExactPlacebo{} to
\PFSTraceEncounterTerraMissingExactRelevant{}; strict success remained
\PFSTraceEncounterTerraMissingStrictPlacebo{}/32. With the positive-only Observation note,
unsupported \supported labels rose from \PFSTraceObservationLunaMissingSupportedPlacebo{}/32 to
\PFSTraceObservationLunaMissingSupportedRelevant{}/32 in Luna's missing-evidence cases and from
\PFSTraceObservationTerraConflictSupportedPlacebo{}/32 to
\PFSTraceObservationTerraConflictSupportedRelevant{}/32 in Terra's conflicting-evidence cases.

\section{Clinical Expert Review Protocols}
\label{app:clinical-review}

\subsection{Earlier Reviews}

Two clinicians judged supplied interpretations for clinical plausibility as chart-search tests:
five lifecycle and five challenge scenarios covering unresolved or conflicting evidence, version
history, Provenance, and medication order--administration links.
The disagreement concerned whether
\texttt{completed} and \texttt{not-done} Immunizations for one claim constitute conflict.
Because interpretations were supplied, this assessed face validity, not independent labels or witnesses.

A separate reviewer reporting emergency-medicine practice received four-state definitions and
one unscored example per state, then 20 shuffled cases with labels and answer key withheld:
five per state and four each for Condition, Immunization, MedicationAdministration,
MedicationRequest, and Observation. The reviewer assigned labels, selected the smallest sufficient
record subset, assessed evidence and search scope, and explained concerns or uncertainty.

All 20 selected record sets matched the program-derived witness. The sole label disagreement
concerned a reference-insufficient Condition with
\texttt{clinicalStatus=resolved} and \texttt{verificationStatus=unconfirmed}: the rule treated it as
neither supporting nor refuting, whereas the reviewer chose conflict and cited the same record.
Concerns occurred in 11 cases, including all five insufficient and four conflicting cases.
Eight concerned evidence clarity, usually requesting clearer value or event displays to distinguish
absence, refutation, and conflict. This single-reviewer audit does not estimate inter-reviewer reliability.

\subsection{Additional Independent Review}
\label{app:review-clinical}

Two additional clinical reviewers assessed the same 24 edited-chart cases from 24 distinct patients,
emphasizing disputed evidence states.
Each of Condition, Immunization, and Observation contributed three insufficient, three conflicting,
one supported, and one refuted case.

\paragraph{Stage A: Independent Labels and Records.}
Reviewers received definitions and examples, but not case-specific reference answers,
the detailed classification policy, or model outputs. The authors report that both completed
Stage A independently before either received Stage B, without sharing answers or reviewing earlier packets.
Agreement rates are reported in Section~\ref{sec:clinical-review}.
Reviewer 1 labeled two reference-refuted cases insufficient, one Condition and one Observation.

The sole clinical-record mismatch was a duplicated selection, retained as entered rather than
inferring a different intended identifier.
Neither reviewer selected the Encounter routing record: full-witness agreement was therefore
0/24 for both. Clinical-record selection and the benchmark's exact-witness endpoint are distinct.

\paragraph{Stage B: Reference Interpretations and Model Answers.}
Reviewers then received the classification rules, reference interpretation, and one actual
model response per case. Responses were selected without inspecting correctness: 12 Luna and
12 Terra, balanced across three instruction conditions and two repeats. This assesses supplied
records and outputs, not full-chart retrieval or population-level model performance.

\begin{table}[t]
\centering
\footnotesize
\setlength{\tabcolsep}{3pt}
\begin{tabular*}{\columnwidth}{@{}L{42mm}@{\extracolsep{\fill}}rr@{}}
\toprule
Judgment & Reviewer 1 & Reviewer 2 \\
\midrule
Reference defensible & 22/1/1 & 24/0/0 \\
Model answer acceptable & 17/6/1 & 19/5/0 \\
Citations adequate, original forms & 17/6/1 & 24/0/0 \\
Citations adequate, clarification & --- & 19/5/0 \\
\bottomrule
\end{tabular*}
\caption{\textbf{Stage B clinical judgments.} Counts are yes/no/uncertain, with 24 judgments
per reviewer. Clarified citations are subsequent author-reported amendments, not an independent review.}
\label{tab:review-clinical-stage-b}
\end{table}

Both rejected the same five model-label errors: two insufficient Conditions predicted supported,
one insufficient Immunization predicted refuted, and two conflicting Observations predicted
supported. Reviewer 1 also rejected a supported Condition reference interpretation and marked
a refuted Condition interpretation uncertain. Other model-answer objections concerned
reference-matching labels.

Reviewer 2's citation clarification changed the five rejected-answer cases from adequate to
unsupported; the original forms are preserved. All categorical and evidence fields were completed,
but all comments were blank, including 12 required case-reviewer explanations, limiting
interpretation of the disagreements. These reviews are not pooled across protocols.

\section{Scoring Definitions, Example, and Rules}
\label{app:worked-examples}

\subsection{Outcome Definitions}

Using the notation of Section~\ref{sec:scoring}, let $R$ contain all chart record IDs and
$P|_C$ denote the cited-record subchart. Denominators in \cref{tab:confirmatory} include all
assigned episodes. Episodes without valid submissions, including failed and turn-limited
primary episodes, score zero on every outcome. For valid submissions:
\begin{align*}
R_C &= \mathbf 1[\text{all IDs in }C\text{ resolve}],\\
K &= \mathbf 1[\text{episode status is completed}],\\
W_C &= \mathbf 1[(\exists W\in\mathcal W:W\subseteq C)\\[-1mm]
    &\qquad{}\lor\ (C\ne\varnothing\ \land\ \operatorname{Exec}(\pi,P|_C).y=y)],\\
Q &= \mathbf 1[\text{all required searches are certified}].
\end{align*}
Here $Q=1$ when no search certificate is required. A failed execution on the restricted chart
does not satisfy the replay alternative in $W_C$. The nonempty-citation condition prevents an empty chart from
justifying an insufficient-evidence label merely because no records remain.
The reported episode outcomes are
\begin{align*}
I_{\mathrm{exact}} &= \mathbf 1[\hat y=y],\\
I_{\mathrm{sufficient}} &= R_C W_C,\\
I_{\mathrm{joint}} &= I_{\mathrm{exact}}I_{\mathrm{sufficient}},\\
I_{\mathrm{search}} &= I_{\mathrm{joint}}Q,\\
I_{\mathrm{strict}} &= K\,\mathbf 1[\hat y=y\land C\in\mathcal W\\[-1mm]
  &\qquad{}\land C\subseteq O\land C\ne R]Q.
\end{align*}

Citation sufficiency is independent of the submitted label.

\subsection{Example Episode: Correct Labels with Incomplete Citations}

The task asks whether the chart contains a hemoglobin result at the most recent completed
follow-up visit and a completed influenza immunization on a specified date. The reference
program selects Encounter Enc; shortened identifiers describe the relevant records below.

\begin{center}
\footnotesize
\begin{tabular}{@{}L{0.25\linewidth}L{0.66\linewidth}@{}}
\toprule
\textbf{Record} & \textbf{Relevant fields} \\
\midrule
Encounter Enc & Most recent completed follow-up visit; status \texttt{finished}. \\
\addlinespace
Observation O1 & Hemoglobin; status \texttt{final}; value absent;
\texttt{dataAbsentReason=unknown}; linked to Enc. \\
\addlinespace
Immunization I1 & Specified influenza vaccine; status \texttt{completed}; requested date. \\
\bottomrule
\end{tabular}
\end{center}

O1 lacks a usable value, and other retrieved hemoglobin values belong to different visits.
Reference and submitted labels therefore agree: insufficient for the primary claim,
supported for the companion claim, and insufficient overall. The declared witness is
$\{\mathrm{Enc},\mathrm{O1},\mathrm{I1}\}$, but the episode cites only I1, omitting the
Encounter needed to determine search scope. Independently, it fails search certification
by retrieving hemoglobin observations without the required Encounter and date restrictions.
Thus exact-answer scoring passes, while evidence sufficiency and strict success fail.

\subsection{Lifecycle Decision Rules}
\label{app:lifecycle-rules}

\begin{table}[H]
\centering
\caption{\textbf{Primary record-level rules.} After code, date, and scope restrictions, Condition
uses \texttt{verificationStatus}; the other rows use \texttt{status}. Other combinations neither
support nor refute. These are benchmark policies for the stated claims.}
\label{tab:lifecycle-rules}
\footnotesize
\setlength{\tabcolsep}{3pt}
\begin{tabular}{@{}L{0.22\linewidth}L{0.45\linewidth}L{0.25\linewidth}@{}}
\toprule
Resource & Supports & Refutes\\
\midrule
Condition & \texttt{confirmed} & \texttt{refuted}\\
Immunization & \texttt{completed} & \texttt{not-done}\\
Observation & \texttt{final}, \texttt{amended}, or \texttt{corrected}, with a nonnull normalized value & \texttt{cancelled}\\
\bottomrule
\end{tabular}
\end{table}

Condition clinical status, including \texttt{resolved}, does not alter the rule. Observation values
may be normalized quantities, scalars, coded values, or component structures. Zero and Boolean false
count as present; no clinical threshold is applied. Cancellation needs no particular
\texttt{dataAbsentReason}.

All inspected primary, replication, four-note, and external snapshots use logical-group rules v3.
For the primary resource types, records join a group through shared nonempty FHIR identifier
\texttt{system}/\texttt{value} pairs within a type, including transitive links. Normalized Observation
components inherit their parent's group. Matching code, date, or Encounter alone does not merge records. Conflict requires support
and refutation within one group. Across groups, conflict takes precedence, then support, refutation,
and insufficient evidence. No selected records also yields insufficient evidence.

\paragraph{Limits of Restricted Replay.}
Restricted replay establishes answer preservation on a reduced chart, not complete search of the
original chart. In a constructed single-claim control, citing only an unrelated Observation preserves
an \insufficient label and passes permissive sufficiency. In the two-claim worked example, retaining
Enc and I1 while removing O1 also passes. Both sets fail exact witness matching; empty citations fail
in both examples. Search certificates independently check required full-chart search responses.

\section{Additional Model Coverage and Execution Settings}
\label{app:review-models}
\label{app:execution-details}
\label{app:gemini-depth3}

The primary Encounter and depth-three studies used \texttt{gpt-5.6-luna}; replication,
four-note, and external studies also used \texttt{gpt-5.6-terra}, through OpenAI Responses
with \texttt{store=false}. Shared defaults are medium reasoning, 2,048 output tokens per request,
12 turns, 24 tool calls, no sampling seed, required sequential tool use, and local histories.
\Cref{tab:review-shared-settings} lists exceptions. Hosted aliases do not identify fixed weights.

\paragraph{Qwen/Luna Interface Comparison.}
The prespecified comparison (Section~\ref{sec:additional-evaluations}) balanced eight patients
per resource family, with 768 replay-verified episodes per model--condition cell. Selection was
independent of outcomes and did not use the designated release test split. Tasks crossed four
states, two widths, and two distractor settings. Qwen used
\texttt{qwen/qwen3.8-27b} through OpenRouter's pinned DeepInfra BF16 route without fallback.
Guidance disclosed only Encounter status.

All 58 terminal failures counted: Luna had four natural and one explicit failure, Qwen 31 and 22.
Intervals use 20,000 patient-bootstrap draws stratified by resource, retaining both repeats and
models. The secondary strict-effect difference, Qwen minus Luna, was $-11.72$ points
(95\% interval $[-16.15,-6.90]$). Missing-evidence label agreement changed by $-19.79$ points
for Luna ($[-27.60,-11.98]$) and $-1.04$ for Qwen ($[-5.21,3.13]$).

\begin{table}[t]
\centering
\footnotesize
\setlength{\tabcolsep}{2pt}
\begin{tabular*}{\columnwidth}{@{}L{31mm}@{\extracolsep{\fill}}L{48mm}@{}}
\toprule
Study/model & Exception to shared defaults \\
\midrule
Qwen interface study & Provider-default parallel calls \\
Claude/Gemini panel & High reasoning; sequential tools \\
Fixed-record diagnostic & No tools; one classification call per case, condition, model, and repeat \\
\bottomrule
\end{tabular*}
\caption{\textbf{Shared settings and study-specific exceptions.} Reasoning labels do not
equalize compute; infrastructure-recovery handling remains study-specific.}
\label{tab:review-shared-settings}
\end{table}

\paragraph{Claude/Gemini Interface Panel.}
The 48-patient panel used 192 units, one per state per patient, balancing width and distractors.
Two repeats under depth-one neutral and depth-two natural/explicit wording yielded 1,152 episodes
per model, 384 per depth-two cell (\cref{tab:review-historical-models}).

\begin{table}[t]
\centering
\footnotesize
\setlength{\tabcolsep}{2pt}
\begin{tabular*}{\columnwidth}{@{}l@{\extracolsep{\fill}}rrr@{}}
\toprule
Model & Natural & Explicit & $\Delta$ [95\% interval] \\
\midrule
Claude Opus 4.8 & 64.06 & 65.63 & $+1.56\;[-2.34,5.21]$ \\
Gemini 3.1 Pro & 18.23 & 14.58 & $-3.65\;[-9.38,2.08]$ \\
\bottomrule
\end{tabular*}
\caption{\textbf{Cross-provider interface comparison.} Gemini is 3.1 Pro Preview.
Rates (\%) average available repeats within unit, units within patient, then patients equally.
Intervals use 20,000 patient-bootstrap draws.}
\label{tab:review-historical-models}
\end{table}

Aliases were \texttt{anthropic/claude-opus-4.8} and \texttt{google/gemini-3.1-pro-preview},
accessed through OpenRouter with a Google Vertex route. These exploratory within-model contrasts
used different cohorts and instruction wrappers and were not run contemporaneously.

Model failures count as failures; unresolved infrastructure outcomes are missing: none for
Claude, three natural and zero explicit depth-two, and four depth-one for Gemini. All 192 units retained an observed
repeat per condition. Recovery amendments allowed one replacement per eligible assignment,
preserving both records.

\paragraph{Partial Gemini Depth-Three Evaluation.}
A completion pass selected counterparts without using correctness, yielding 88 pairs from six
patients, 44 per width and all four evidence states. Five patients contributed 16 pairs each, one
eight. Only DiagnosticReport and Procedure paths are represented. \Cref{tab:confirmatory-width}
weights patients equally within width.

Gemini 3.1 Pro Preview used Vertex Priority, shared medium-reasoning limits, zero-data retention,
and no fallback. Static inputs and settings matched within pairs, but requests were
noncontemporaneous and collection controls changed. This incomplete, availability-selected sample
supports descriptive results only, without confidence intervals or significance tests.

Replay verified responses and scores. All 22 no-answer episodes (12 natural, ten explicit)
remain failures. Exact answers numbered 59/88 versus 56/88, but required search checks failed in
54/59 and 52/56 of those episodes, explaining most of the gap from strict success.

\section{Retrieval, Search, and Citation Diagnostics}
\label{app:review-diagnostics}
\label{app:diagnostics}

\subsection{Paired Retrieval and Label Transitions}

This follow-up analysis includes all pairs whose primary reference label is \insufficient,
retaining failures and primary-study response repairs. Retrieval requires the target Encounter in a successful tool
response, regardless of search certification; its ID alone does not suffice
(\cref{tab:retrieval-label-transitions}).

% Verified against anonymous_aggregates.json; reference-insufficient pairs only.
\begin{table}[t]
\centering
\caption{\textbf{Encounter retrieval and primary-label changes.}
Arrows run from natural to explicit wording. The last column counts newly returned Encounters
among label regressions. Primary: 225 patients; replication: two repeats on 48 patients.}
\label{tab:retrieval-label-transitions}
\footnotesize
\renewcommand{\arraystretch}{1.08}
\setlength{\tabcolsep}{1.8pt}
\begin{tabular*}{\columnwidth}{@{}l@{\extracolsep{\fill}}rrrr@{}}
\toprule
\shortstack[l]{Cohort /\\model} & Pairs & \shortstack[r]{Encounter\\returned} &
\shortstack[r]{Correct\\primary label} & \shortstack[r]{New retrieval /\\regressions} \\
\midrule
Primary Luna & 900 & $661\to900$ & $443\to264$ & $170/226$ \\
Rep.\ Luna & 384 & $266\to384$ & $214\to163$ & $66/82$ \\
Rep.\ Terra & 384 & $186\to384$ & $283\to180$ & $112/125$ \\
\bottomrule
\end{tabular*}
\end{table}

\paragraph{Retrieval-Conditioned Missing-Evidence Labels.}
Among retrieved episodes, correct labels numbered 208/661 versus 264/900 for primary Luna,
104/266 versus 163/384 for replication Luna, and 95/186 versus 180/384 for Terra.
Restricting instead to pairs retrieved in both arms gave changes of $-1.36$ points
(95\% interval $[-4.68,1.83]$), $+4.89$ ($[-0.39,9.96]$), and $+3.23$ ($[-1.07,7.11]$),
respectively, using 10,000 patient-cluster bootstrap draws. Neither comparison is causal:
guidance changes retrieval, and the same visit need not yield the same downstream evidence.

The 226 primary regressions comprised 91 supported Condition labels, 40 refuted and 13 conflicting
Immunization labels, and 78 supported and four refuted Observation labels. Condition was the largest
group overall, but Observation accounted for 55/56 regressions with retrieval in both arms. These counts
do not establish how valueless final Observations were interpreted.

\subsection{Follow-Up Search-Scoring Sensitivity}
This earlier-cohort analysis was not prespecified. Reference IDs were returned despite failed
search checks in 415 primary-Luna, 172 replication-Luna, and 22 Terra episodes; identical IDs
do not establish equivalent scope. Complete same-snapshot pagination chains (26 primary and
eight replication-Luna candidates) added no strict successes. Verified equivalent target-code
and same-Encounter-ID queries converted 1,301 failures to strict successes. The guidance effect
changed from $+10.56$ to $+17.67$ points for primary Luna, $+11.13$ to $+23.63$ for replication
Luna, and $+44.53$ to $+44.34$ for Terra.

The Qwen/Luna study prespecified program-equivalence and reference-scope-coverage
sensitivities. The latter gave Qwen 348/768 versus 361/768 successes, $+1.69$ points
(95\% interval $[-1.30,4.69]$). Fixed-snapshot coverage is not semantic query equivalence.

\subsection{Fixed-Record Classification and Citations}
For 48 previously studied synthetic patients (16 per rule family), identical complete search
outputs covered four states under generic, routing-only, or routing-plus-full-policy instructions.
Two models and two repeats yielded 2,304 responses. The full policy discloses the evaluator's rules
and adds length, so their effects are not separated.

\begin{table}[t]
\centering
\footnotesize
\setlength{\tabcolsep}{2pt}
\begin{tabular*}{\columnwidth}{@{}ll@{\extracolsep{\fill}}rrr@{}}
\toprule
Model & Instructions & Clinical & Routing & ID error \\
\midrule
Luna & Generic & 23 & 0 & 0 \\
Luna & Routing & 22 & 0 & 0 \\
Luna & Full policy & 0 & 13 & 1 \\
Terra & Generic & 18 & 9 & 0 \\
Terra & Routing & 21 & 9 & 0 \\
Terra & Full policy & 0 & 328 & 0 \\
\bottomrule
\end{tabular*}
\caption{\textbf{Fixed-record citation failures.} Each row has 384 responses. Mutually exclusive
categories are clinical-only omission, routing-only omission, or an unobserved truncated ID
with a missing clinical record. No response added extra observed records or duplicate citations.
These are not the original tool-use error proportions.}
\label{tab:review-fixed-citations}
\end{table}

Both models matched all 384 labels with the complete policy, but exact-witness agreement
was 370/384 for Luna and 56/384 for Terra because routing omissions still failed the
unchanged citation endpoint.

\end{document}